\pdfoutput=1

\documentclass[11pt]{article}

\usepackage{acl}

\usepackage{times}
\usepackage{latexsym}

\usepackage[shortlabels]{enumitem}
\usepackage{amsmath}
\usepackage{scalerel,graphicx,xparse}
\usepackage{pifont}
\usepackage{float}
\usepackage{amsfonts}
\usepackage{adjustbox}
\usepackage{bbm}

\usepackage[T1]{fontenc}

\usepackage[utf8]{inputenc}

\usepackage{microtype}

\usepackage{inconsolata}

\usepackage{booktabs}
\usepackage{graphicx}
\usepackage{multirow}
\usepackage{tabularx}
\usepackage{framed}

\definecolor{bottlegreen}{rgb}{0.0,0.42,0.31}
\definecolor{donorred}{RGB}{228.,116.,95.}
\definecolor{reciepientblue}{RGB}{0,152,251}

\usepackage[normalem]{ulem}

\newcommand{\khealth}[0]{K Health}
\newcommand{\mh}[0]{\textit{Must Have}}
\newcommand{\nth}[0]{\textit{Nice to Have}}
\newcommand{\kqa}[0]{\textsc{K-QA}}
\newcommand{\fscore}[0]{FActScore}

\newcommand{\example}[1]{\emph{``#1''}}
\newcommand{\numofcurated}[0]{201}
\newcommand{\numofraw}[0]{1,212}
\newcommand{\checked}{\ding{51}}
\newcommand{\crossed}{\ding{55}}

\title{\kqa{}: A Real-World Medical Q\&A Benchmark}

\author{
    Itay Manes$^1$, 
    Naama Ronn$^1$,
    David Cohen$^1$, 
    Ran Ilan Ber$^1$, \\
    {\bf Zehavi Horowitz-Kugler$^1$}, 
    {\bf Gabriel Stanovsky$^{2}$} \\
    $^1$K Health Inc, New York, NY \\
    $^2$School of Computer Science, The Hebrew University of Jerusalem \\ 
    {\texttt{\{itay.manes,david.cohen,ran.ilanber\}@khealth.com}}
    }

\begin{document}
\maketitle
\begin{abstract}
Ensuring the accuracy of responses provided by large language models (LLMs) is crucial, particularly in clinical settings where incorrect information may directly impact patient health. To address this challenge, we construct \kqa{}, a dataset containing \numofraw{} patient questions originating from real-world conversations held on K Health (an AI-driven clinical platform). We employ a panel of in-house physicians to answer and manually decompose a subset of \kqa{} into self-contained statements. Additionally, we formulate two NLI-based evaluation metrics approximating recall and precision: 
(1) \emph{comprehensiveness}, measuring the percentage of essential clinical information in the generated answer and 
(2) \emph{hallucination rate}, measuring the number of statements from the physician-curated response contradicted by the LLM answer.
Finally, we use \kqa{} along with these metrics to evaluate several state-of-the-art models, as well as the effect of in-context learning and medically-oriented augmented retrieval schemes developed by the authors.
Our findings indicate that in-context learning improves the comprehensiveness of the models, and augmented retrieval is effective in reducing hallucinations. 
We make \kqa{} available to to the community to spur research into medically accurate NLP applications. \footnote{The data and the evaluation script are available at \url{https://github.com/Itaymanes/K-QA}. Results and model comparisons can be viewed at \url{https://huggingface.co/spaces/Itaykhealth/K-QA}}

\end{abstract}

\begin{figure*}[t]
  \centering
  \includegraphics[width=\textwidth]{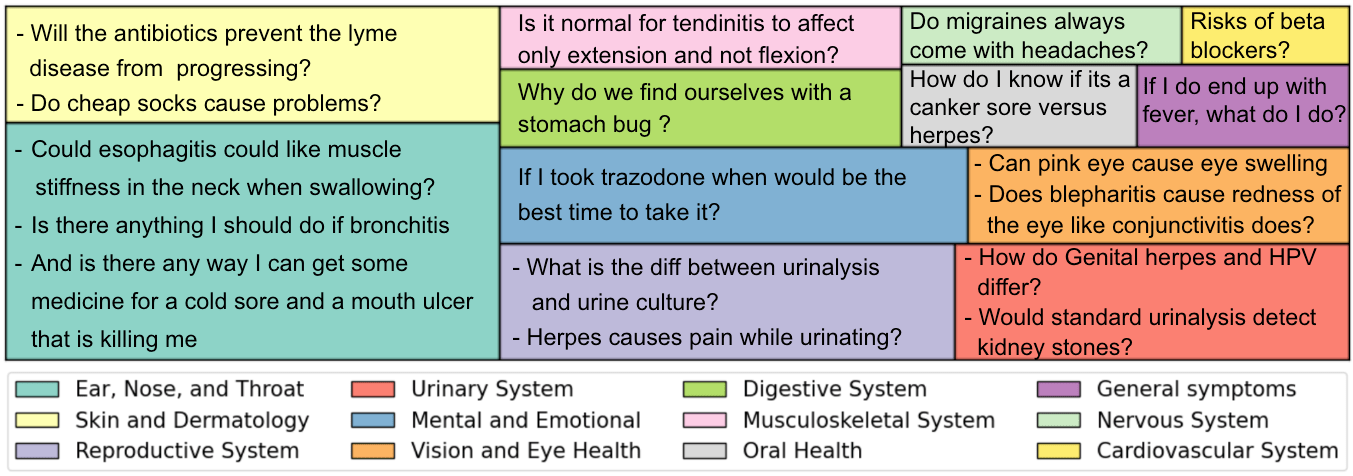}
  
  \caption{
  Visualization of \kqa{}, with box sizes indicating the distribution of patients' reported chief complaints across a wide range of healthcare topics. The questions are open-ended and diverse.}
  \label{fig:data_viz}
\end{figure*}

\section{Introduction}
Recent advancements in large language models (LLMs) have led to a growing interest in their application in the medical domain in patient-facing applications, where LLMs hold the promise of providing laypersons with high-quality advice at a relatively low cost~\citep{medpalm,han2023medalpaca}. For instance, in response to the question \example{What's good for muscular pain?}, a good patient-facing response may include, in addition to medical information (e.g., a muscle relaxant), also the advice \example{Seek medical attention if you have numbness or tingling in limbs}.

However, there is a lack of benchmarks reflecting user needs and corresponding medically-accurate answers to test these models under real-world conditions. 
Most existing benchmarks assume textbook questions with multiple-choice or span-based answers~\citep{tsatsaronis2015overview,LiveMedQA2017,jin2019pubmedqa}.
In contrast, real-world questions, like \example{Is there any way I can get some medicine for cold sore and ulcer that is killing me?}, can often include various interacting medical conditions (\example{cold sore}, \example{ulcer}), use ambiguous, non-medical jargon (\example{that is killing me}), and require long-form, nuanced answers.

In this work, we present \kqa{}, a medical QA benchmark containing
\numofraw{} deidentified questions asked by real users on \khealth{},\footnote{https://khealth.com} an AI-driven clinical platform with
over 8 million unique users.
The questions in \kqa{} were curated from \khealth{}'s vast database of patient-physician interactions, aiming to capture stand-alone medical questions.  These can be answered solely based on the information provided in the question, and do not require any prior knowledge about the patient's history or demographics.
The resulting corpus is diverse and challenging, spanning over
100 different medical conditions (see examples in Figure~\ref{fig:data_viz}).

To evaluate state-of-the-art models against \kqa{}, a team of 12 in-house medical doctors rigorously answered \numofcurated{} questions from the dataset in a free-text format.
Doctors consulted credible medical sources, such as UpToDate\footnote{https://www.wolterskluwer.com/en/solutions/uptodate} and PubMed\footnote{https://pubmed.ncbi.nlm.nih.gov/} to provide accurate and scientifically-backed answers. Their answers were further reviewed by an experienced overseeing physician and tested for inter-annotator agreement.

To allow fine-grained evaluation, doctors decomposed each answer into an average of roughly 8 minimal semantic content units~\citep{nenkova2007pyramid}, resulting in over 1.5K individual statements. In addition, the importance of each statement was manually marked as either (1) \mh{}, indicating that a model must include this statement in order to be medically accurate (e.g., providing all contraindications for a drug) , or (2) \nth{}, indicating the statement is supplemental in nature (e.g., providing additional conditions where this drug may be helpful). 

Following recent work on evaluation of text generation, we use the decomposed ground-truth answers in a natural language inference (NLI)-based evaluation of predicted answers~\citep{honovich-etal-2021-q2,laban-etal-2022-summac,aharoni-etal-2023-multilingual}. Concretely, we define two complementing evaluation metrics. First, \emph{comprehensiveness}, which is similar to recall, and measures the percentage of ground-truth statement included in the predicted answer. In order to excel in this metric, a model must cover as many of the \mh{} statements annotated by doctors. Second, \emph{hallucination rate}, which is similar to precision, and measures how many of all ground-truth statements \emph{contradict} the predicted answer. To excel in this metric models must not produce \emph{any} medically-inaccurate statements. We find that recent LLMs, like GPT-4, are able to approximate such metrics with good performance, nearing human agreement on the task.

Finally, we evaluate various state-of-the-art LLM-based architectures on \kqa{}, spanning a wide range of families, including open- and closed- source models, zero-shot and in-context learning prompting, and retrieval-augmented generations. We find that all models struggle on \emph{comprehensiveness}, with the best performing model covering only 67.7\% of medically-important statements, and while hallucinations seem to decrease with model size and augmented generation, all models may still provide advice which may be medically dangerous in subtle ways, which may not be noticeable by every-day users.

We hope that future work adopts \kqa{} and accompanying metrics as a valuable benchmark to produce medically-accurate NLP applications which can be deployed in real-world scenarios.

\section{Background}
\label{sec:related_work}
Our study is closely related  to two topics: long-form question answering, particularly within the medical domain, and the evaluation of factuality in generated text. In this section, we provide a review of prior work in these fields.  In addition, we also highlight key features of \kqa{} and contrast it with previous work in Table \ref{tab:datasets}.

\begin{table*}[tb!]
\small
\centering
\begin{tabularx}{\textwidth}{@{}p{\dimexpr4cm+1\tabcolsep} *{5}{>{\centering\arraybackslash}X} l@{}}

\toprule
Dataset &
  \begin{tabular}[c]{@{}c@{}}Consumer\\ Health\end{tabular} &
  \multicolumn{1}{c}{\begin{tabular}[c]{@{}c@{}}Open\\ Domain\end{tabular}} &
  \multicolumn{1}{c}{\begin{tabular}[c]{@{}c@{}}Patient-Physician \\ Interaction\end{tabular}} &
  \multicolumn{1}{c}{\begin{tabular}[c]{@{}c@{}}Answer\\Decomposition\end{tabular}} &
  \multicolumn{1}{c}{\begin{tabular}[c]{@{}c@{}}Answer\\ Format\end{tabular}} &
  \\ \midrule

\begin{tabular}[c]{@{}l@{}}BioASQ \citep{tsatsaronis2015overview} \end{tabular} & \multicolumn{1}{c}{\crossed} & \multicolumn{1}{c}{\crossed} & \multicolumn{1}{c}{\crossed} & \multicolumn{1}{c}{\crossed}  & \multicolumn{1}{c}{\begin{tabular}[c]{@{}c@{}}span-based \\ 
 \& binary\end{tabular}}
\\ \midrule

\begin{tabular}[c]{@{}l@{}}MedQA \citep{jin2021disease}\end{tabular} & \multicolumn{1}{c}{\crossed} & \multicolumn{1}{c}{\checked} & \multicolumn{1}{c}{\crossed} & \multicolumn{1}{c}{\crossed}  &   \multicolumn{1}{c}{\begin{tabular}[c]{@{}c@{}}multiple \\ choice\end{tabular}}  \\ \midrule

\begin{tabular}[c]{@{}l@{}}LiveMedQA \citep{LiveMedQA2017}\end{tabular}  & \multicolumn{1}{c}{\checked} & \multicolumn{1}{c}{\crossed} & \multicolumn{1}{c}{\crossed}  &  \multicolumn{1}{c}{\checked} & \multicolumn{1}{c}{retrieval}   \\ \midrule

\begin{tabular}[c]{@{}l@{}}MedicationQA \citep{abacha2017overview}\end{tabular} 
& \multicolumn{1}{c}{\checked} & \multicolumn{1}{c}{\crossed} & \multicolumn{1}{c}{\crossed}  & \multicolumn{1}{c}{\crossed} & \multicolumn{1}{c}{retrieval}\\ \midrule

\begin{tabular}[c]{@{}l@{}}MedQuAD \cite{BenAbacha:MEDINFO19} \end{tabular} 
& \multicolumn{1}{c}{\checked} & \multicolumn{1}{c}{\crossed} & \multicolumn{1}{c}{\crossed}  & \multicolumn{1}{c}{\crossed} & \multicolumn{1}{c}{retrieval} \\ \midrule

\begin{tabular}[c]{@{}l@{}}MEDIQA-AnS \citep{savery2020question}\end{tabular}    & \multicolumn{1}{c}{\checked} &   \multicolumn{1}{c}{\crossed}  & \multicolumn{1}{c}{\crossed}   &  \multicolumn{1}{c}{\crossed}   & \multicolumn{1}{c}{ranking}   \\ \midrule

\begin{tabular}[c]{@{}l@{}}HealthQA \citep{singhal2023large}\end{tabular}    & \multicolumn{1}{c}{\checked} &   \multicolumn{1}{c}{\checked}  & \multicolumn{1}{c}{\crossed}   &  \multicolumn{1}{c}{\crossed}   & \multicolumn{1}{c}{-}   \\ \midrule

\textbf{\kqa{} (Ours)}   & \multicolumn{1}{c}{\checked} & \multicolumn{1}{c}{\checked} & \multicolumn{1}{c}{\checked} & \multicolumn{1}{c}{\checked} &  \multicolumn{1}{c}{\begin{tabular}[c]{@{}c@{}}long-form \\ generation \end{tabular}}   \\ 
\bottomrule
\end{tabularx}

\caption{Comparison between \kqa{} and previous benchmarks in the field of medical question-answering. A \checked{} or \crossed{} indicates the presence or absence of a particular feature in a dataset. For a more comprehensive discussion, please refer to the main text.
}

\label{tab:datasets}
\end{table*}

\subsection{Medical QA Benchmarks}
Several diverse health-related question-answering datasets have been compiled, including over biomedical scientific literature~\citep{tsatsaronis2015overview, jin2019pubmedqa} and medical examinations~\citep{zhang2018medical, pmlr-v174-pal22a}.
The majority of these datasets rely on multiple-choice or span extraction
~\citep{jin2022biomedical}, which simplify the evaluation process but do not reflect complexity of free-form responses which are often needed in real-world situations~\citep{gehrmann2023repairing}.

In the context of consumer health questions, our dataset is different from existing benchmarks like MEDIQA-AnS \citep{savery2020question}, LiveMedQA \citep{LiveMedQA2017} and MedicationQA \citep{BenAbacha:MEDINFO19} in several additional key ways.  While these datasets source their questions from users searching healthcare websites via the ChiQA system \citep{demner2020consumer} and retrieve answers through keyword matching, ours originates from authentic patient-physician interactions, ensuring genuine medical inquiries. 
Furthermore, our dataset includes free-form open-domain responses carefully curated by medical professionals.  In addition, the answers in \kqa{} are segmented into finer atomic statements, enabling fine-grained evaluation. 
\subsection{Factuality Evaluation in Long-Form Generation}
In addition to the \kqa{} benchmark, we introduce an evaluation suite to compare the ability of different models to generate
factual and coherent text, covering all of the required medical information. 
Our evaluation framework is inspired by recent work, which decomposes long text into atomic facts and verifies their factuality using NLI models~\citep{min2023factscore,kamoi2023wice,chen2022propsegment}. Alongside examining whether the generated text entails ground-truth statements, we also introduce novel metrics that model differences in the importance of these statements, classifying each as \mh{} or \nth{}.

\subsection{Challenging LLM Benchmarks}
Our work joins a recent line work which produces test sets which are challenging for state-of-the-art LLMs. For example, the Bamboogle benchmark consists of 125 multi-hop questions which stump popular search engines~\citep{Press2022MeasuringAN}, while the GPQA benchmark contains 445 graduate-level questions in various domains~\citep{Rein2023GPQAAG}. \kqa{} consists of \numofraw{} questions, as well as a subset of 201 answers, specially curated by in-house physicians.

\begin{figure*}[tb!]
  \centering
  \includegraphics[width=\textwidth]{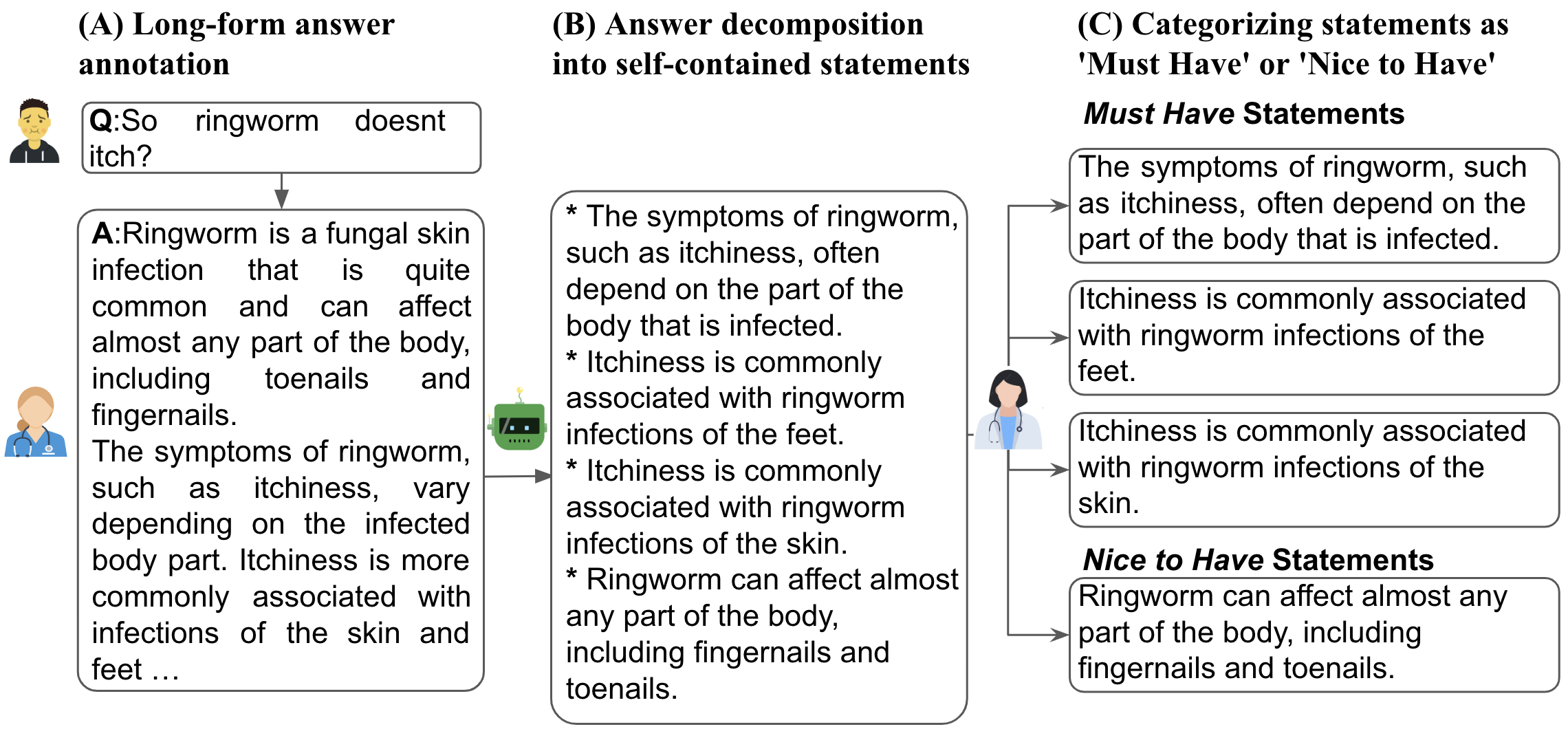}
  \caption{
  High-level description of the \kqa{} dataset creation process, starting with a physician's considered offline response to an actual patient query obtained from patient-physician interactions. We then utilized LLMs to break-down the response into self-contained statements, subsequently categorized by a panel of medical experts as \textit{\mh{}} or \textit{\nth{}}. The example was simplified for presentation purposes. 
  }
  \label{fig:example}
\end{figure*}

\section{The \kqa{} Benchmark}

\label{sec:datasets}
In this section, we describe the process of curating and annotating the \kqa{} dataset, depicted in Figure~\ref{fig:example}. \kqa{} consists of two portions - a medium-scale corpus of diverse real-world medical inquiries written by patients on an online platform (Section~\ref{sec:large-scale}) and a subset of rigorous and granular answers, annotated by a team of in-house medical experts (Section~\ref{sec:curation}). Finally, in Section \ref{sec:stats}, we present an analysis of the dataset, illustrating its diversity in medical characteristics.


\subsection{Curating Questions from Real-World Patient-Physician Conversations} 
\label{sec:large-scale}
All of the questions in \kqa{} originate from de-identified real-world text-based conversations held on \khealth{}.
These conversations contain a wide variety of user intents, such as billing inquiries or prescription renewals, alongside a wealth of queries on varied medical subjects (see Figure~\ref{fig:data_viz}).  


Our goal in creating \kqa{} is to extract from this large and noisy corpus a diverse dataset of \emph{medical} questions which can be used to test automated models' ability to provide factual and comprehensive medical answers. In particular, we aim for the extracted questions to be as \emph{stand-alone} as possible, without relying on the patient's medical record or the context of the medical discourse.
For example (adapted from Figure~\ref{fig:data_viz}), \kqa{} includes questions such as \example{How do Genital herpes and HPV differ?}, while we omit questions such as \example{Can this allergic reaction be related to my age?} which assumes prior knowledge about the patient and their previous symptoms. 

To achieve this, we performed a rigorous manual annotation, aided by a preliminary automatic preprocessing step. First,  we used an open-source BERT-based classifier \cite{devlin2018bert}, fine-tuned for distinguishing questions from statements, such as \example{sounds like hives to me}.\footnote{\url{https://huggingface.co/mrsinghania/asr-question-detection}} Next, we applied regular expressions to filter questions about logistics (e.g., billing or delivery instructions). 
This preprocessing yielded roughly 26K questions, each individually assessed by a medical professional to identify those suitable as stand-alone questions.
Overall, this resulted in a dataset of \numofraw{} diverse questions, each accompanied by a physician's classification of the medical condition discussed in the question, according to ICD-10 conventions (e.g., the question in Figure~\ref{fig:data_viz} is classified as \example{Dermatitis, unspecified}) \cite{world1993icd}. 
\subsection{Annotating Granular Physician Answers}
\label{sec:curation}
We provide comprehensive and granular answers for a diverse subset of \kqa{} questions, annotated in three stages by a team of 12
in-house medical doctors. This subset enables us to automatically compare different LLMs against high-quality expert answers. 

\begin{figure}[h]
\includegraphics[width=\columnwidth]{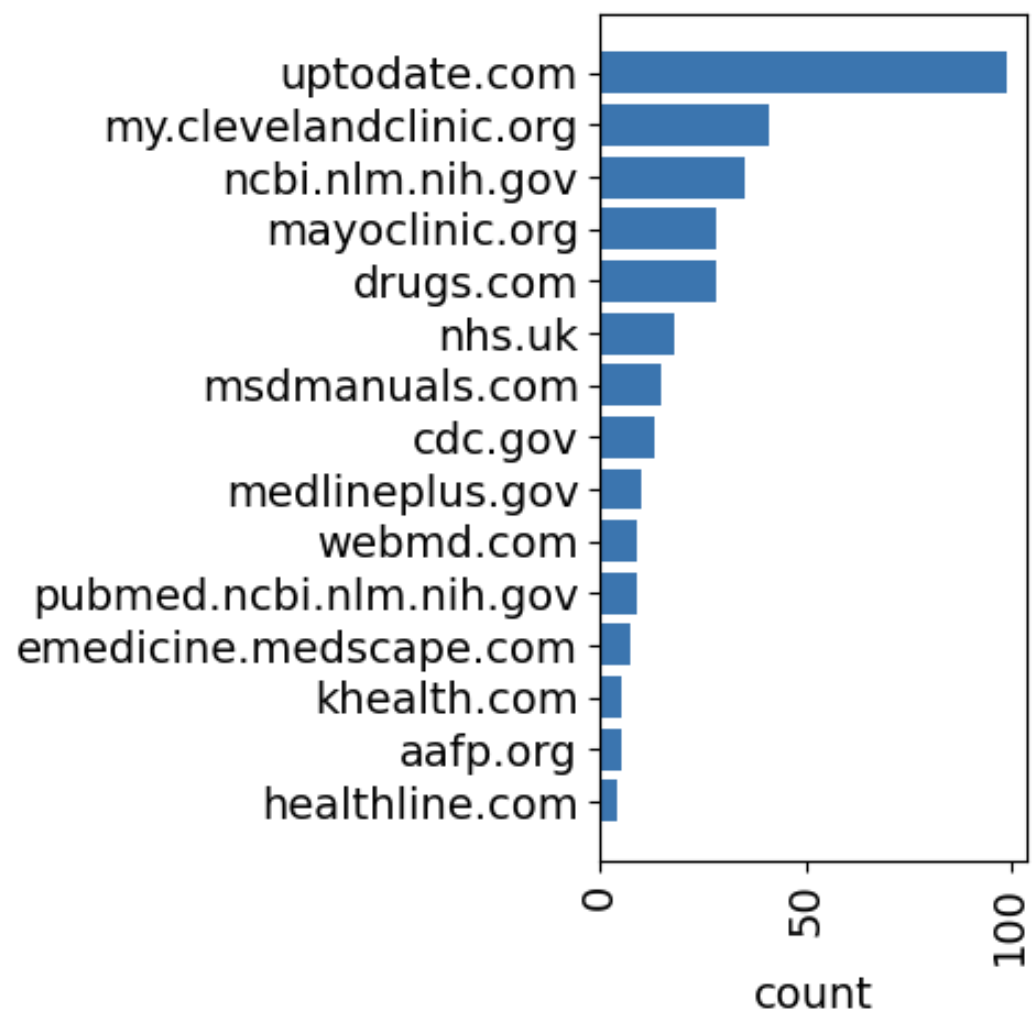}
\caption{The 15 most used medical resources by the annotators during the curation of the long-form answers.}
\label{fig:ann_resources}
\end{figure}

\paragraph{Step 1: Long-form answer annotation.}
In the first annotation step, exemplified in Figure~\ref{fig:example}(A), 
six medical physicians were tasked with providing free-form responses to different sets of questions from \kqa{}, while an additional physician reviewed their answers and advised where needed. Overall, the first step required roughly 400 skilled person hours (at a cost of roughly 26K USD, based on average physician hourly pay in the U.S.), during which 201 questions from \kqa{}  were answered.
Each physician was granted unlimited time and access to reputable medical resources such as UpToDate and PubMed for referencing purposes (refer to Figure~\ref{fig:ann_resources} for the most used resources). Notably, they were explicitly instructed to \emph{avoid} using any generative language models or services.
To best emulate the requirements from a user-facing model in the medical domain, annotators were further instructed to write answers tailored for a lay audience seeking consumer-health information. 
For example, note how the answer in Figure~\ref{fig:example} regarding ringworm strays from medical jargon.
\paragraph{Step 2: Answer decomposition into self-contained statements.} \label{step-2}
Following literature on the evaluation of text generation via minimal semantic content units~\citep{nenkova2007pyramid,liu-2022-low}, we guided annotators to decompose answers into self-contained statements. Each statement is expected to capture a distinct fact and include sufficient context for independent evaluation.

Answer decomposition is presented in Figure~\ref{fig:example}(B), illustrating the decomposition of a natural answer into atomic statements. 
A more complex example of medically oriented statement decomposition is presented below. In response to a question about treating hypertension in diabetic patients, the physician recommends ACE inhibitors or ARBs, either alone or in combination with other drugs like calcium channel blockers and thiazides. This scenario illustrates an exclusive OR relation often observed in medical contexts, where multiple treatments are optional but not advised together. 
To address the dependency between treatment options, we use \nth{} to enable us to preserve the physician's intention, emphasizing the importance of taking either ACE or ARBs and suggesting an additional optional treatment for each. This results in one \mh{} statement:
\example{A recommended treatment includes either ACE or ARBs, but not both.}, and two \nth{} statements: (1) \example{ARBs can be taken alone or with other medications, such as calcium channel blockers and thiazides.}; and (2) \example{ACE can be taken alone or with other medications, such as calcium channel blockers and thiazides.} The full annotation guidelines and more examples are provided in Appendix~\ref{app:deco-guidelines}.

This step was carried out by a panel of 6 medical  doctors (distinct from the annotators in the first step) who deconstructed each answer into individual statements. To assist in this process, the panel utilized GPT-4 with a few-shot prompt suggesting potential answer decompositions. The full prompt is provided in Appendix \ref{app:decom-prompt}.
The annotators could amend or remove noisy statements, as well as add any missing statements, which they did for 6.86\% of the automatically generated statements. In total, this process yielded 1,589 annotated statements, averaging roughly 7.9 statements per answer. The completion of this phase required a total of approximately 30 hours at the cost of approximately 2K USD.
\paragraph{Step 3: Categorizing statements as \mh{} or \nth{}.}\label{step-3}
At the last step, we asked the same group of medical professionals from step 2 to classify each statement into one of two categories: (1) \mh{} – facets of the answer which are crucial to convey to a patient when providing medical advice; or (2) \nth{} -- statements which are supplemental or informative, but not clinically crucial.  For example, as can be seen in Figure~\ref{fig:example}(C), the long-form answer regarding the itchiness of ringworm was decomposed into four statements, three of which were deemed as \mh, while a statement which provided  information about ringworm which is not related to its itchiness was deemed as \nth{}. 
This process was carried out collaboratively, facilitating discussions within the medical group to collectively assess and reach consensus regarding the perceived levels of importance, amounting to approximately 20 person hours, at the cost of roughly 1.5K USD. 

The categorization into \mh{} and \nth{} represents a discrete approach to assigning importance to statements. However, in a broader context, this method can be extended to assign various weighted scores to each statement and each metric.

\begin{figure*}[tb!]
  \centering
  \includegraphics[width=\textwidth]{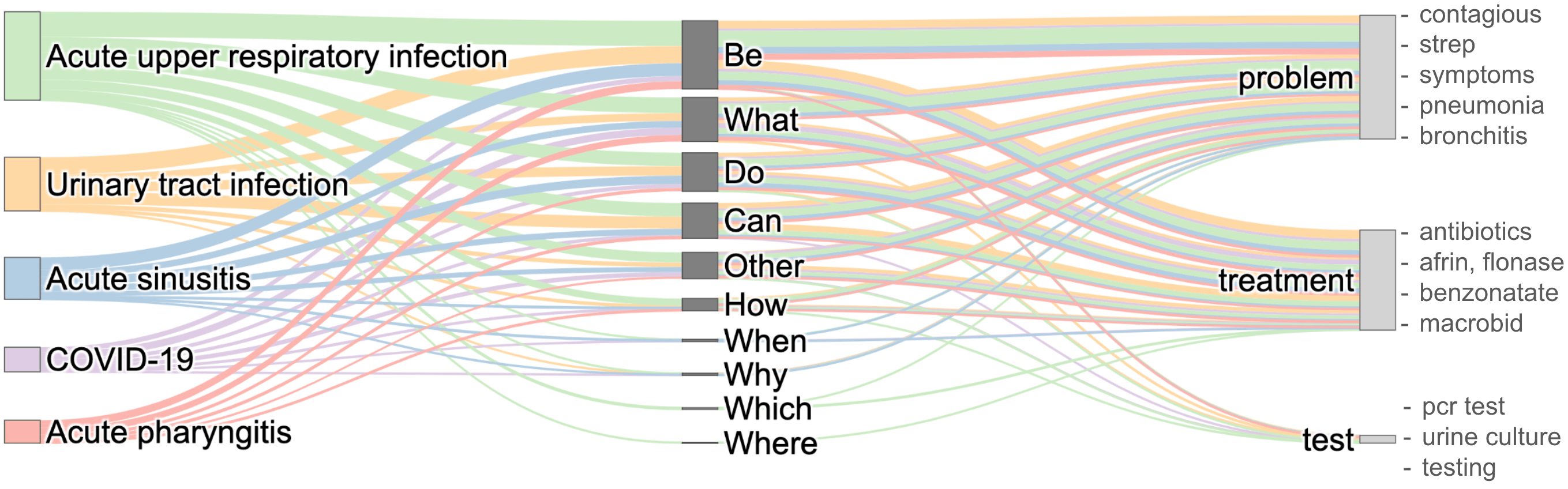}
  
  \caption{
    Distribution of the top 5 most prevalent medical conditions, the types of questions related to each condition, and the frequencies of clinical entities within those questions. On the far right, the text most frequently matched with the clinical entities is displayed.
  }
  \label{fig:sankey}
\end{figure*}

\label{sec:datasets_analysis}
\subsection{Dataset Statistics}
\label{sec:stats}
\begin{table}[tb!]
\centering
\begin{tabular}{@{}lll@{}}
\toprule
          & \multicolumn{1}{c}{Count} & \multicolumn{1}{c}{\# Words (avg.)} \\ \midrule
Questions & \multicolumn{1}{c}{\numofraw}   & \multicolumn{1}{c}{10.06}           \\
Answers   & \multicolumn{1}{c}{201}   & \multicolumn{1}{c}{88.52}         \\
Statements \\
$\quad$ \textit{\mh}    & \multicolumn{1}{c}{892}  & \multicolumn{1}{c}{14.9} \\
$\quad$ \textit{\nth}   & \multicolumn{1}{c}{697}  & \multicolumn{1}{c}{13.74} 
\\ \bottomrule
\end{tabular}
\caption{Statistics of the \kqa{} benchmark.
}
\label{tab:vocab}
\end{table}

Our \kqa{} dataset is derived from a diverse group of 1,055 unique users featuring \numofraw{} questions, including 201 answers meticulously curated by physicians. Table \ref{tab:vocab} shows detailed statistics on statements and word counts, while information on the distribution of age and biological sex among users can be found in Table \ref{tab:age-sex}. 

\begin{table}[tb!]
  \centering
  \begin{tabular}{ccrr}
    \toprule
    \multirow{2}{*}{Age Group} & \multicolumn{2}{c}{Sex (\% of users)} \\ \cmidrule(lr){2-3}
    & Female & Male \\ 
    \midrule
    18-25 & 9.09  & 6.67  \\
    26-45 & 36.97  & 30.30  \\
    46-60 & 9.70  & 3.64  \\
    60+ & 2.42  & 1.21  \\
    \bottomrule
  \end{tabular} 
  \caption{Distribution of the users in \kqa{} by age group and biological sex.}
  \label{tab:age-sex}
\end{table}

The questions in our dataset were written to address a wide array of health concerns, as evidenced in Figure \ref{fig:data_viz}. These patients discuss 172 different medical conditions, coded according to the ICD-10 system. 
The diversity in questions is highlighted in Figure \ref{fig:sankey}, focusing on the top five most prevalent diagnoses and their distributions across various 
question types, such as Be, WH-questions and other forms.
Additionally, on the right side of Figure \ref{fig:sankey}, frequencies of clinical categories (\textit{problem}, \textit{treatment}, and \textit{test}) are displayed, classified by using an  open-source, fine-tuned named entity recognition BERT-based model.\footnote{\url{https://huggingface.co/samrawal/bert-base-uncased_clinical-ner}} 
For example, a question like \example{Is it usually normal for someone to have side effects when first starting vitamins?} is labeled with both \textit{problem} (``side effects'') and \textit{treatment} (``vitamins''). 


\begin{figure*}[tb!]
  \centering
  \includegraphics[width=\textwidth]{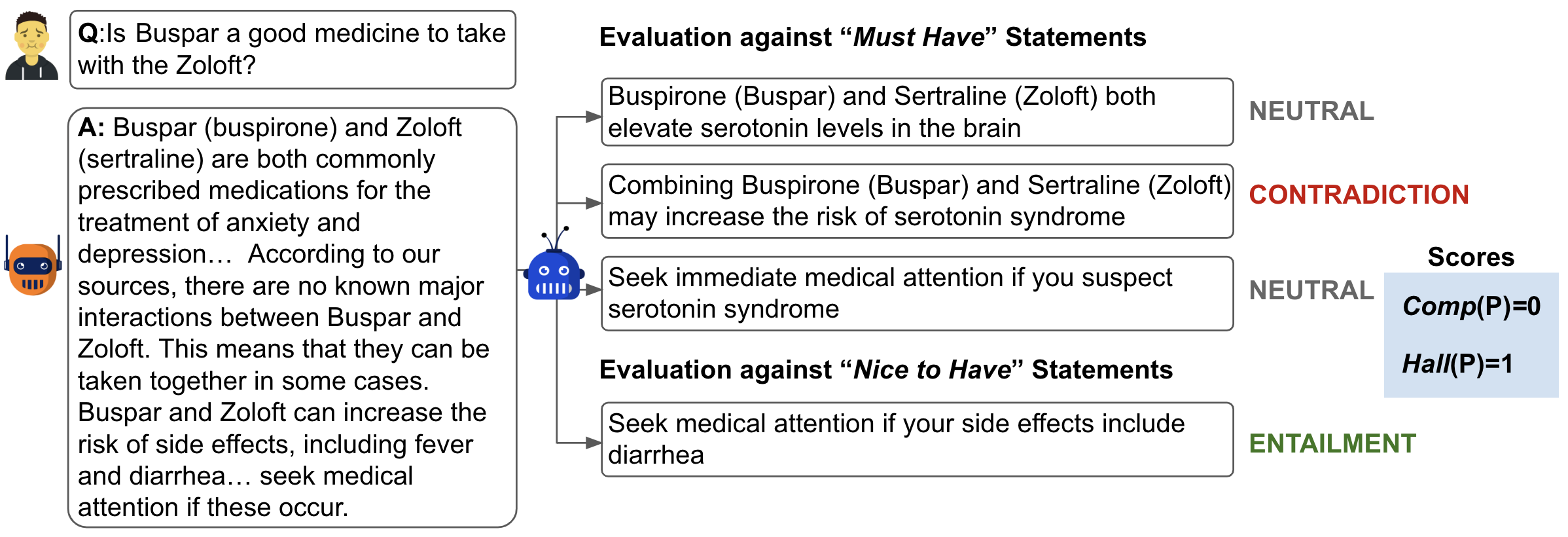}
  \caption{
    An example of the evaluation procedure begins with a patient's question and a generated answer from a language model. 
    Each statement is then tested separately and automatically with a model in an NLI framework to determine its relationship to the generated answer. Finally, the metrics are computed, where $Hall(\hat{P})$ is equal to 1 since there is only one contradiction, and $Comp(\hat{P})$ is equal to $0$ because none of the \mh{} statements were entailed. Different robot symbols signify different models, and the example was simplified for presentation. 
  }
  \label{fig:eval_example}
\end{figure*}


\section{Evaluation Metrics for \kqa{}}
\label{sec:eval}
To evaluate models against our \kqa{} benchmark, we propose a natural language inference~(NLI; \citealp{Dagan2005ThePR,Bowman2015ALA}) evaluation framework, following recent text generation evaluation~\citep{honovich-etal-2021-q2,laban-etal-2022-summac,aharoni-etal-2023-multilingual}. 

Our evaluation framework is inspired by \fscore{} \citep{min2023factscore}, a metric that measures factual \emph{precision} by computing the percentage of \emph{atomic facts} in a generated answer supported by a reliable external source. 
Unlike \fscore{}, which automatically generates statements and assigns them equal importance, our approach involves predefined medical statements with varying clinical significance. This modification enables us to extend the framework and establish a proxy metric for factual \emph{recall} as well.

We consider a predicted answer as a \emph{premise} and each ground-truth statement derived from an annotated answer as an \emph{hypothesis}. Intuitively, a correctly predicted answer should entail every ground-truth statement. This formulation aims to quantify the extent to which the model's answer captures the semantic meaning of the gold answer, abstracting over the wording chosen by a particular expert annotator.
 
As formulated below, we devise two NLI-based metrics:  \emph{comprehensiveness} and \emph{hallucination rate}. 
These adapt the evaluation of text generation to the medical domain by taking into account \kqa{}'s annotation of \textit{\mh{}}, i.e., clinically crucial facets of information, and \textit{\nth{}} statements, which are supplemental in nature. 
Both metrics were aggregated across all assessed questions, where higher values of the comprehensiveness metric and lower values of hallucination rates  indicate better performance.
Figure \ref{fig:eval_example} provides an example illustrating the complete process of evaluating a generated answer and deriving these metrics.

Formally, let $\hat{P}$ denote the model's predicted answer,  \textit{\mh{}} represents the set of ground-truth statements marked as crucial, \textit{\nth{}} represents the set of ground-truth statements marked as supplemental, and by $S = \textit{\mh{}} \cup \textit{\nth{}}$ is the set of all statements in the gold reference answer.

\paragraph{Comprehensiveness metric.}
This metric measures how many of the clinically crucial claims are included in the predicted answer.
\begin{equation}
    Comp(\hat{P}) = \frac{|\{x \in Must\_Have | \text{$\hat{P}$ entails $x$\}}|}{|Must\_Have|}
\end{equation}  

I.e., similarly to recall, $Comp(\hat{P}) \in [0, 1]$ measures how many ground-truth statements were conveyed in the predicted answer. We particularly focus on those statements marked as crucial by medical experts and which do not penalize models for not covering supplemental statements, as these may be somewhat arbitrary. 

\paragraph{Hallucination rate.}
This metric measures how many of the ground-truth statements contradict the model's answer.
\begin{equation}
    Hall(\hat{P}) = |\{x \in S | \text{$\hat{P}$ contradicts $x$\}}|
\end{equation}  

I.e., $Hall(\hat{P}) \in \{0,1, ..., | S | \}$ penalizes answers that contradict any of the statements and hence discourages models from making any sort of false medical statements. Similar to precision, a model can achieve a perfect score trivially by generating an empty answer $\hat{P} = \emptyset$ since, by definition, no hypothesis contradicts an empty premise. 


\paragraph{Automatic evaluation.} 
Following work on NLI-based evaluation, we approximate the metrics above via an automated NLI model. We employed GPT-4 as the language model in conjunction with few-shot Chain-of-Thought (CoT; \citealp{wei2022chain}), a prompt that generates sequential intermediary text representations.
To assess the quality of the evaluation framework, we randomly selected 50 pairs of questions and their corresponding generated answers from the models described in Section \ref{sec:experiments}. This process yielded 402 unique statements pertaining to the specified set of 50 questions.

Three physicians received instructions on how to classify the logical relationship for each triplet (\emph{question}, \emph{answer}, \emph{statement}) into one of three NLI categories. The inter-agreement among annotators was assessed using Fleiss’ kappa ({$\kappa$; \citealp{fleiss1971measuring}) and pairwise agreement. For the three human annotators, the pairwise agreement was $83.2\%$, and the $\kappa$ was calculated to be 0.70, signifying moderate to substantial agreement among raters.
The agreement with the majority vote of the annotators and the automated model was $83.0\%$, indicating that the model can perform at a level comparable to human annotators for this complex task.


\section{Evaluating State-of-the Art Models}
Following the creation of \kqa{} and the formulation of evaluation metrics, we turn to
evaluate the current state of the art in this challenging task.
\subsection{Experimental Setup}
\label{sec:experiments}

\paragraph{Models.} We use \kqa{} to evaluate the medical capabilities of 7 recent LLM-based models from diverse families and model sizes. Specifically, we evaluate two 7B instruction-tuned open access models: Mistral~\citep{jiang2023mistral},  and MedAlpaca~\citep{han2023medalpaca} which was built upon LLaMA \cite{touvron2023llama} and trained specifically for biomedical tasks, three recent closed instruction-tuned LLMs: Open AI's GPT-3.5 and GPT-4~\citep{brown2020language, openai2023gpt4}, and Google's PALM-2~\citep{anil2023palm}, and finally two recent commercial closed generation search engines: BARD,\footnote{https://bard.google.com/} and Bing Chat.\footnote{https://www.bing.com}
We use zero temperature sampling for all models, except for BARD and Bing Chat, which do not allow setting temperature.

\paragraph{Retrieval augmented generation (RAG).}
We note that BARD and Bing Chat differ from the other 5 models in our evaluation in that they can reportedly augment their prompt with content retrieved from the external sources, albeit in an undisclosed manner. To examine the effect that retrieved content may have on the performance of the other models, we implement Retrieval Augmented Generation approach~(RAG; \citealp{lewis2020retrieval}), which produces responses by conditioning the language model on both the input query and retrieved content.
To achieve this, we index publicly available medical documents aimed at the lay audience (such as MayoClinic\footnote{https://www.mayoclinic.org/} and NHS\footnote{https://www.nhs.uk/}) aiming for medical-specific RAG.
All the documents in this RAG are publicly available, which is distinct from the primary sources that the physician annotators used to create their answers (Figure \ref{fig:ann_resources}).

\paragraph{Prompts.}
Most of our evaluations use the same vanilla zero-shot prompt without prompt engineering which only presents the question, without any additional instructions.
In addition, for some models we also report results on another empirically engineered prompt which includes three in-context examples, to
explore some of the effect that in context learning (ICL) may have on performance.
We note however that finding optimal prompts for a given task and model is an open question ~\cite{liu2023pre}, which we leave in our case as interesting avenue for future work.

\subsection{Results}
\label{sec:results}
The results for all models are shown in Table \ref{tab:results}, in terms of the comprehensiveness and hallucination rate metrics defined in Section~\ref{sec:eval}. We anlayze additional facets of the models' performance in Table~\ref{tab:res_stats}.
Below, we highlight key observations based on these results.

\begin{table}[tb!]
\centering

\normalsize
\adjustbox{max width=\columnwidth}{

\newcolumntype{C}[1]{>{\centering\arraybackslash}p{#1}}
\begin{tabular}{lcccc}
\toprule
Model &  \emph{Comp} $\uparrow$  &  \emph{Hall} $\downarrow$ & \% respond  & 
\\ \midrule

MedAlpaca 7B & 31.4 & 56.7 & 100\\
Mistral 7B   &  47.6   & 28.4 & 100 \\
PALM-2     &  50.8 &  31.3  & 100\\ 
BARD  & 62.5   &  28.4  & 95.0\\      
Bing Chat  &  57.3 & 25.9 & 99.5 \\  
GPT-3.5   &  56.2   & 27.9  & 100 \\
GPT-3.5+ICL    &  59.5   & 23.4 & 99.5 \\
GPT-3.5+RAG   &  50.5   & 17.9  & 89.0 \\
GPT-3.5+ICL+RAG    & 62.9  &  \textbf{15.4} & 96.0\\
GPT-4      &  57.5  & 23.9  & 100\\   
GPT-4+ICL      &  \textbf{67.7} & 25.4 & 100 \\   
GPT-4+RAG     & 52.2  &  22.9  & 91.5 \\
GPT-4+ICL+RAG     & 65.2  &  24.4 & 100 \\
\bottomrule


\end{tabular}%
}
\caption{Comparing models, where ICL represents the addition of three in-context examples, and RAG is the medical retrieval augmented setup, as detailed in Section \ref{sec:experiments}. 
The performance of the highest scoring model is \textbf{bolded} for each metric. \% \emph{respond} indicates the percentage of generations that do not abstain from answering the questions.}
\label{tab:results}
\end{table}

\paragraph{Attaining high comprehensiveness is challenging even for state-of-the-art models.}
Across all models and prompts, the comprehensiveness metric (\emph{Comp}) consistently remains below 68\%.  This is evident even in cases with longer texts, as seen in the BARD model, which contains nearly three times as many words per answer (242.1, Table \ref{tab:res_stats}) compared to the physician's response (88.36, Table \ref{tab:vocab}).
This underscores the models' difficulty in capturing what physicians consider critically important. 
Additionally, ICL improves comprehensiveness by instructing the model to include elements beyond a direct response to the question, such as assuming underlying medical concerns in patient inquiries.
Some of the low scores may be explained by model's abstaining from answering certain questions, as shown in Table~\ref{tab:results}.

\paragraph{While hallucinations seem rare, they could potentially lead to unintended and unsafe medical recommendations.} 
The best hallucination rate, achieved by GPT-3.5+ICL+RAG, represents a contradiction of roughly 30 statements out of 1500 annotated examples. However, some of the hallucinations may lead to subtle yet dangerous advice. For example, in Figure~\ref{fig:eval_example}, the physician's statement asserts that \example{Combining Buspar and Zoloft may increase the risk of serotonin syndrome}, while, on the other hand, the model claims that \example{there are no known major interactions between Buspar and Zoloft}. This contradiction can impact the patient's medication intake. We note that finding cause of error in such cases is hard, and physicians are also prone to making  dangerous errors.
This particular error can be attributed to a combination of missing information within publicly available medical sources and by the LLM assuming that their omission implies the drug combination is safe.

\paragraph{For the GPT models in our evaluation, it seems that larger models lead to improved comprehensiveness…}  In Table \ref{tab:results}, we observe that GPT-4 outperforms GPT-3.5 under every comparable setting. These findings align with those of \citet{roberts2020much}, who demonstrated, in the task of open-domain question answering, that the capability of a pretrained model to answer questions without access to any external knowledge scales with the model's size.

\paragraph{… yet the larger the GPT model, the more it seems to introduce new hallucinations.} Revisiting Table \ref{tab:results}, we see that the improved comprehensiveness comes at the cost of an increase in the hallucination rate. 

\paragraph{Domain-specific RAG reduces hallucinations.}
Among all configurations, GPT-3.5+ICL+RAG demonstrates the fewest hallucinations while maintaining a comparatively good comprehensiveness score. Upon close examination, we found that the tradeoff with comprehensiveness is partly due to its tendency to abstain from answering in certain questions (see \emph{\% respond} column in Table~\ref{tab:res_stats}). We note that this may actually be desired over misinformation in a patient-facing application.
For this model, when computing the metrics only over the answered questions, it receives a \emph{Comp} score of 65.5\% and a \emph{Hall} score of 16.1, which is still the lowest in hallucinations but with the second-highest comprehensiveness.
However, Bing Chat and BARD, which also abstain occasionally (e.g., \example{I’m unable to help, as I am only a language model and don’t have the ability to process and understand that.}), appear to underperform compared to their base models, even though they use some variants of GPT and PALM.
This discrepancy might stem from our prompts lacking task optimization and their generic web retrieval, especially failing to focus on consumer health inquiries in the medical domain from reliable sources.



\paragraph{MedAlpaca performs poorly on \kqa{}.}
Even though MedAlpaca was fine-tuned specifically for the biomedical domain and intended for use as medical conversational AI, it exhibites the poorest results on both metrics, with an especially high hallucination rate. These findings indicate a mismatch between closed-QA (e.g., medical exams and short answers) and real-world patient questions which require the generation of long medical answers.

\begin{table}[tb!]
\centering

\normalsize
\adjustbox{max width=\textwidth}{

\newcolumntype{C}[1]{>{\centering\arraybackslash}p{#1}}
\begin{tabular}{lcccc}
\toprule
Model &  Use search &  \# words &
\\ \midrule
        MedAlpaca 7B &  {\texttimes} & 48.6 \\
        Mistral 7B & {\texttimes}  & 68.4 \\
        PALM-2 & {\texttimes}  & 80.0 \\
        BARD & {\checkmark}  & 242.1 \\
        Bing Chat & {\checkmark}  & 95.7 \\
        GPT-3.5 & {\texttimes} & 60.8 \\
        GPT-3.5+ICL & {\texttimes} & 80.3 \\
        GPT-3.5+RAG & {\checkmark} & 96.3 \\
        GPT-3.5+ICL+RAG & {\checkmark} & 173.2 \\
        GPT-4 & {\texttimes} & 63.1 \\
        GPT-4+ICL & {\texttimes} & 132.8 \\
        GPT-4+RAG & {\checkmark} & 89.8 \\
        GPT-4+ICL+RAG & {\checkmark} & 130.1 \\
\bottomrule

\end{tabular}%
}
\caption{Statistics of the generated responses across models. \# \emph{words} represents the average number of words in a response}
\label{tab:res_stats}
\end{table}


\section{Conclusion}
\label{sec:Conclusion}
In this work, we introduce \kqa{}, a question-answering benchmark that includes real-world patients' questions and carefully curated fine-grained answers by physicians. In addition, we formulate metrics that quantify how well a predicted answer covers important information and to what extent it contradicts gold answers. Our evaluation shows that while models improve with size and augmented generation, there is still a lot of room for improvement, especially in the comprehensiveness of the answers and ensuring that they do not include medically inaccurate statements.

\section{Ethics Statement}
The data in \kqa{} originates from deidentified real-world patient conversations that have been manually reviewed to ensure there are no leaks. In particular, we verified that the questions do not disclose any private medical information, and revolve around general medical questions. The answers in \kqa{} were manually written by medical doctors, who did not use any automated writing assistance and wrote their answers with a general audience in mind. Our legal team has reviewed and approved the methodology used.

\section{Limitations}
One of the major limitations of our evaluation approach is its reliance on LLMs for approximating the entailment relation between ground-truth and predicted answers. While this was done in various recent works, it may propagate noise into the evaluation process, and yield a costly evaluation protocol. To mitigate this concern, we measure the agreement between human annotators and predicted labels, finding overall good agreement, while reducing evaluation costs an important avenue for future work~\citep{Perlitz2023EfficientB}.

\section{Acknowledgments} 
We would like to express our gratitude to Carmel Vider, Lee Herzog, Tamar Brufman, Nadia Mordenfeld, Lior Hayat, Ohad Simons, Jessica Thien, Shira Ben Porat, Itay Omer, Assaf Frajman, and Noa Bleistein for curating and reviewing the medical answers. 
In addition, we would like to thank Nadav Rabani and Ya'ara Arkin for insightful discussion that influenced the direction of this work.
\bibliography{anthology,custom}

\appendix
\section{General scoring framework}\label{app:math}
In order to expand upon the definitions provided in \ref{sec:eval}, we define $w\in\mathcal{W}$ as the weight assigned to a specific statement $s\in\mathcal{S}$ for a given question. Additionally, we consider an NLI model $\mathcal{N}(premise, hypothesis)$ designed to classify each pair of \textit{(premise, hypothesis)} into one of three labels: $l \in \{entail, contradict, neutral\}$. Within this framework, we view the hypothesis as the statement, and the generated response ($\hat{P}$) as the premise, denoted as $\mathcal{N}(\hat{P}, s)$. We define $f(\mathcal{N}(\hat{P}, s), l)$ as a function that takes the output of the NLI model and a predefined logical relation, such as "does not contradict," and returns a boolean value. The formula representing this process is as follows:
\begin{align*}
    \sum_{s \in \mathcal{S}} w(s) \cdot f(\mathcal{N}(\hat{P}, s), l)
\end{align*}
This mathematical expression quantifies how well the generated response aligns with a predefined logical condition, while taking into account the weights assigned to individual statement. 
The formulation of this equation is aligned with the metrics presented at section \ref{sec:eval}. For \emph{Hall} computation, $w(s)$ is set to 1, whereas for \emph{Comp}, $w(s)$ takes the value of 1 if $s$ is in \mh{} and 0 if $s$ falls within \nth{}. 

\section{Annotators' guidelines}
\subsection{Decomposition to Self-Contained Statements} \label{app:deco-guidelines}
We aim to evaluate the medical accuracy of responses generated by language models, specifically concerning stand-alone questions within medical conversations initiated by users. Given the potential for these answers to be open-ended, the evaluation task presents inherent challenges. To address this, we've devised a set of guidelines that break down answers into two components: "Must-have" statements, deemed essential for inclusion, and "Nice-to-have" statements, which, while beneficial, are not obligatory. Our objective is to formulate statements that are concise, accurate, definitive, and self-contained. It is imperative to ensure that the curated statements are medically correct and logically well-structured. The evaluation process is conducted independently for each statement, emphasizing the importance of avoiding overlap between statements to maintain clarity. 
\subsection*{Guidelines for Various Scenarios}
Below, a specific explanation is provided for different scenarios, accompanied by examples of good and bad options for decomposing an answer.
\begin{itemize}[left=0.3em, itemsep=0ex]
    \item \textbf{List of Unrelated Crucial Entities:} If the answer comprises a list of entities (e.g., red flags, vaccines, symptoms), and each entity is independently significant, consider treating each entity as a separate and distinct statement. However, if the list of entities is not critical to include in its entirety (e.g., suggestions for weekly menu options), these entities should be combined into a single statement and categorized as \nth{}. The validation for these statements should ensure non-contradiction with the physician's input.
\end{itemize}

\begin{table}[H]
\scriptsize
\centering
\begin{tabularx}{0.85\columnwidth}{|X|X|X|}
\hline
\textbf{Question} & \textbf{\mh{} - Good} & \textbf{\mh{} - Bad} \\
\hline
I am a young healthy adult, flying to Brazil next month. What vaccination should I take? & -Vaccination for Yellow Fever is recommended before traveling to Brazil. 

-Vaccination for Typhoid is recommended before traveling to Brazil. & Since you are traveling to Brazil next month, it is recommended to get fully vaccinated for several vaccines, including Typhoid and yellow fever. \\
\hline
\end{tabularx}
\end{table}

Why is this example considered suboptimal?
\begin{itemize}[label={-}]
    \item Single Answer Instead of Two: Instead of presenting two separate statements, the response combines both vaccines into a single answer. To enhance clarity and evaluation, it is recommended to break down such responses into distinct and independent statements, especially when the mentioned vaccines are not interdependent.
    \item Excessive Length: The response is too long and contains unnecessary information, particularly with the inclusion of prefixes like "Since you…". The focus should be on keeping the information concise and relevant.
    \item Overly Specific: The mention of the timeframe "next month" is overly specific and potentially misleading. It is crucial to provide information that is essential and directly related to the question.
\end{itemize}

\begin{itemize}[left=0.3em, itemsep=0ex]
    \item \textbf{AND/OR statements:} When entities have a logical relationship (e.g., treatment options), express them in statements following their logical connection rather than separating them. Entity 1 AND/OR Entity 2. 
\end{itemize}

\begin{table}[H]
\centering
\scriptsize
\begin{tabularx}{0.85\columnwidth}{|X|X|X|}
\hline
\textbf{Question} & \textbf{\mh{} - Good} & \textbf{\mh{} - Bad} \\
\hline
What is the best hypertension treatment for patients who are also diabetic? & Either angiotensin-converting enzyme (ACE) inhibitors OR angiotensin receptor blockers (ARBs), but not both. & -Angiotensin-converting enzyme (ACE) inhibitors

-Angiotensin receptor blockers (ARBs) \\

\hline
 & \textbf{\nth{} - Good} & \textbf{\nth{} - Bad} \\
\hline
 &  -angiotensin receptor blockers (ARBs) can be taken alone or with the following medications (thiazide and/or ccb). 

-angiotensin-converting enzyme (ACE) can be taken alone or with the following medications (thiazide and/or ccb)
& \\
\hline
\end{tabularx}
\end{table}

Why is this example considered suboptimal?
\begin{itemize}[label={-}]
    \item Misleading statements: The entities ACE and the ARB are dependent on each other. If the language model (LLM) provides an answer recommending the patient take both ACE and ARB, it would be medically incorrect but might receive a high score in our evaluation method. In such cases, as ACE and ARB are distinct treatment options, we need to combine them in a statement to emphasize that only one of them can be prescribed, not both.
    \item Lack of inclusivity: Besides ACEs and ARBs, the treatment plan may involve other medications. We want to ensure that the response does not contradict the LLM's answer.
\end{itemize}
Moreover, if there are additional medications that can be prescribed alongside ACE or ARB, they should be listed in parentheses next to thiazide and/or CCB.

\begin{itemize}[left=0.3em, itemsep=0ex]
    \item \textbf{IF Statements:} Similar to AND/OR statements, maintain the coherence of "IF statements" if the cause is not a stand-alone factor.
\end{itemize}

\begin{table}[H]
\centering
\scriptsize
\begin{tabularx}{0.85\columnwidth}{|X|X|X|}
\hline
& \textbf{\nth{} - Good} & \textbf{\nth{} - Bad} \\
\hline
What is the best hypertension treatment for patients who are also diabetic? & If lifestyle modifications alone are not effective in reducing blood pressure, medications may be necessary; & -Medications are needed to reduce the blood pressure.

-Sometimes life modifications are not enough to reduce the blood pressure. \\
\hline
\end{tabularx}
\end{table}
Why is this example considered suboptimal?
\begin{itemize}[label={-}]
    \item Misleading statements: Splitting the IF statement leads to misleading statements. Even if the "bad examples" are medically correct, we might deviate from the intended verification of the IF statement.
\end{itemize}

\begin{itemize}[left=0.3em, itemsep=0ex]
    \item \textbf{Drugs Inclusion:} Include both the family drug name (generic) and trade name when dealing with drugs. For instance, "Low-risk drugs during pregnancy include aminosalicylates, such as sulfasalazine (Azulfidine) and mesalamine (Asacol, Pentasa).
\end{itemize}

\section{Annotators' interface}
Figures \ref{taggers_framework} display the annotation interface used for human evaluation during dataset creation. In this interface, annotators executed the second and third steps (described in \ref{step-2} \& \ref{step-3}). They were presented with the patient question, the free-form answer written by the physician, and the suggested decomposition of statements provided by GPT-4. Annotators were tasked with confirming, modifying, or removing the suggested decomposition to ensure relevance and self-containment. Additionally, they categorized the statement into one of the categories: \mh{}, \nth{}, or deemed it irrelevant.

\clearpage 
\begin{figure}[h]
\includegraphics[width=\textwidth]{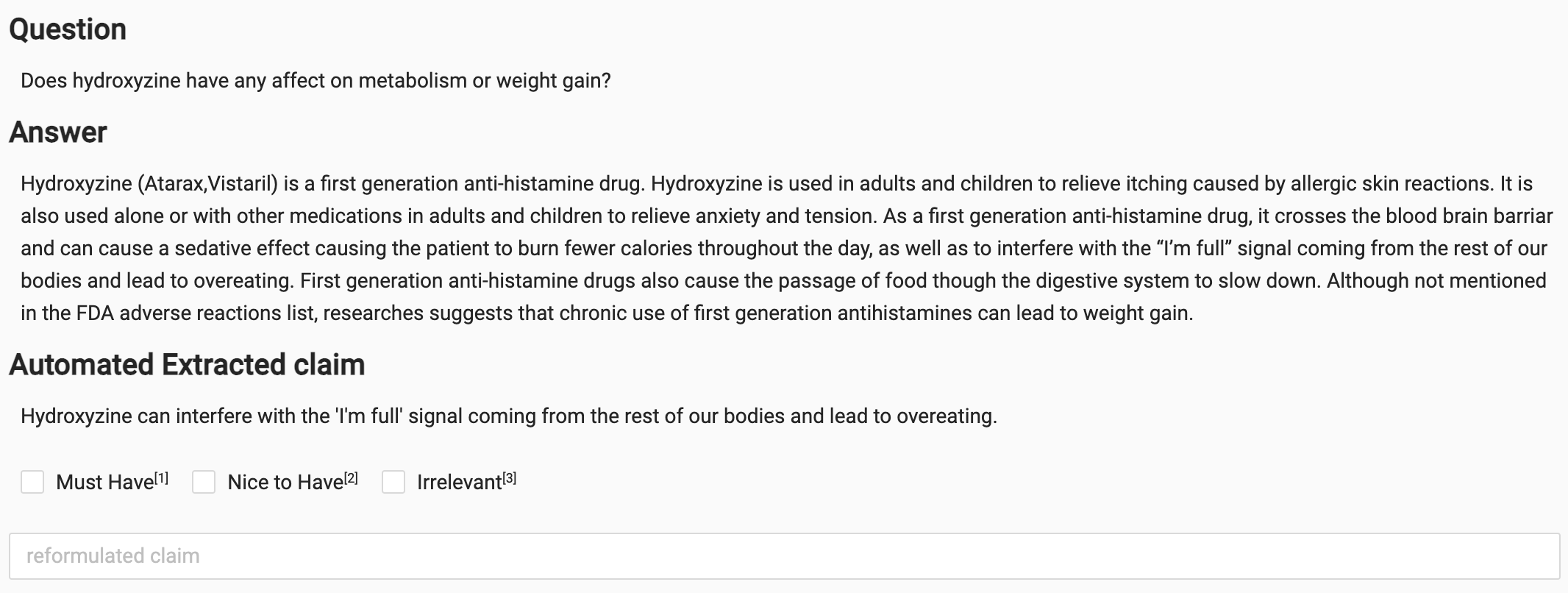}
\caption{An example of the annotator's interface.}
\label{taggers_framework}
\end{figure}

\clearpage 



\section{Prompt templates}
\subsection{Decomposition Free-Form to Statements} \label{app:decom-prompt}
\framebox{
  \scriptsize
  \ttfamily
  \parbox{\textwidth}{
    \# OVERALL INSTRUCTIONS \\
    You are an expert in understanding logical relationships.
    This is a Semantic Content Unit (SCU) extraction task. Given a pair of Question and Answer, your goal is to create a list of self-contained and concise claims. Each claim should be able to stand alone and be independent of other claims.
    Your claims should encompass all the information present in the answer. \\
    
    \# TASK INSTRUCTIONS \\
    - List of Possible Causes: For scenarios involving multiple entities like red flags, vaccines, symptoms, etc., generate separate claims for each entity. This increases the number of claims. \\
    - OR Claims:  When medical entities are presented in an "OR" context, treat them as distinct claims. \\
    - IF Claims: When an "if statement" is present, preserve the "if statement" context while creating the claim. \\
    - XOR Claims: When entities have an XOR logical relationship (e.g., treatment options), create separate claims for each option. \\
    
    \# EXAMPLE CLAIM FORMAT
    - List Format: "Possible cause for [CONDITION] in [DEMOGRAPHIC] can be [ENTITY]." \\
    - OR Format: "Possible causes include: [ENTITY X], [ENTITY Y], and [ENTITY Z]." \\
    - OR Format: "The [CONTEXT] of treatments such as [TREATMENT X], [TREATMENT Y], and [TREATMENT Z], is not well established."
    - IF Format: "[CONTEXT], please seek medical attention if [CONDITIONS]." \\ 
    - XOR Format: "Either take [TREATMENT X] or [TREATMENT Y], but not both." \\
    -----\\
    \{format\_instructions\} \\
    ------\\
    
    \# TASK EXAMPLE \\
    Question: I am a 33-year-old female with right lower abdominal pain, what could it be?
    Answer: Possible causes for right lower abdominal pain in a young female are Appendicitis, Inflammatory bowel disease, Diverticulitis, Kidney stone, urinary tract infection, Ovarian cyst or torsion, Ectopic pregnancy, Pelvic inflammatory disease, endometriosis.
    Please seek medical attention if the pain is sudden and severe, does not go away, or gets worse, is accompanied by fever, nausea and vomiting, or if you have noticed blood in urine or in stool. \\
    Claims: [
    Possible cause for right lower abdominal pain in a young female: Appendicitis,
    Possible cause for right lower abdominal pain in a young female: Ovarian cyst or torsion,
    Possible cause for right lower abdominal pain in a young female: Ectopic pregnancy,
    Possible cause for right lower abdominal pain in a young female: Pelvic inflammatory disease,
    Possible cause for right lower abdominal pain in a young female: Kidney stone,
    Possible cause for right lower abdominal pain in a young female: Urinary tract infection,
    Possible cause for right lower abdominal pain in a young female: Diverticulitis,
    Possible cause for right lower abdominal pain in a young female: Inflammatory bowel disease,
    Possible cause for right lower abdominal pain in a young female: Endometriosis,
    Please seek medical attention if the pain is sudden and severe,
    Please seek medical attention if the pain is accompanied by fever,
    Please seek medical attention if the pain is accompanied by nausea and vomiting,
    Please seek medical attention if the pain is accompanied by blood in urine,
    Please seek medical attention if the pain is accompanied by blood in stool,
    Possible cause for right lower abdominal pain in a young female: Emotional stress
    ]\\
    
    \# TASK EXAMPLE \\
    Question: So what does the non reactive mean for the hep a igm
    Answer: Hep A IgM refers to a specific type of antibody called Immunoglobulin M (IgM) against the virus hepatitis A. When infected with hepatitis A, these antibodies are detectable at symptom onset and remain detectable for approximately three to six months. These antibodies might also be detectable in the first month after hepatitis A vaccination.
    A negative or non-reactive result means no IgM antibodies against hepatitis A found in your serum, meaning the absence of an acute or recent hepatitis A virus infection. \\
    Claims: [
    A negative or non-reactive result means that there were no IgM antibodies against hepatitis A found in your serum,
    The absence of IgM antibodies against hepatitis A in your serum indicates the absence of an acute or recent hepatitis A virus infection,
    Hep A IgM refers to a specific type of antibodies called Immunoglobulin M (IgM) against the virus hepatitis A,
    These antibodies might also be detectable in the first month after hepatitis A vaccination,
    These antibodies remain detectable for approximately three to six months after infection,
    When infected with hepatitis A, these antibodies are detectable at the time of symptom onset
    ]\\
    
    \# TASK EXAMPLE \\
    Question: What medications are contraindicated for a pregnant woman with ulcerative colitis?
    Answer: methotrexate (Otrexup, Rasuvo, RediTrex) and thalidomide (Contergan, Thalomid) are both considered contraindicated for treatment of UC in pregnancy. possible treatment for UC during pregnancy include low-risk drugs such as aminosalicylates (sulfasalazine and mesalamine), immunomodulators (azathioprine, cyclosporine A ,6-mercaptopurine) and corticosteroids. Biological agents such as Infliximabl, Adalimumab, Vedolizumab and Ustekinumab is generally avoided during pregnancy as their safety in pregnancy is not well established yet. \\
    Claims: [
    Methotrexate (Otrexup, Rasuvo, RediTrex) is contraindicated for treatment of ulcerative colitis in pregnancy,
    Thalidomide (Contergan, Thalomid) is contraindicated for treatment of ulcerative colitis in pregnancy,
    Aminosalicylates (sulfasalazine and mesalamine) are considered low-risk drugs for treatment of ulcerative colitis during pregnancy,
    Immunomodulators (azathioprine, cyclosporine A, 6-mercaptopurine) are considered low-risk drugs for treatment of ulcerative colitis during pregnancy,
    Corticosteroids are considered low-risk drugs for treatment of ulcerative colitis during pregnancy,
    Treatment for ulcerative colitis during pregnancy with biological agents such as Adalimumab is generally avoided during pregnancy as their safety in pregnancy is not well established yet,
    Treatment for ulcerative colitis during pregnancy with biological agents such as Vedolizumab is generally avoided during pregnancy as their safety in pregnancy is not well established yet,
    Treatment for ulcerative colitis during pregnancy with biological agents such as Infliximab is generally avoided during pregnancy as their safety in pregnancy is not well established yet,
    Treatment for ulcerative colitis during pregnancy with biological agents such as Ustekinumab is generally avoided during pregnancy as their safety in pregnancy is not well established yet,
    ] \\ 
    
    \# YOUR TASK\\
    Question: \{question\} \\
    Answer: \{answer\} \\
    Claims:
    
  }
}
\clearpage

\subsection{Automatic Evaluation}
\framebox{
  \scriptsize
  \ttfamily
  \parbox{\textwidth}{
    \# OVERALL INSTRUCTIONS \\
    - You have a deep understanding of logical relationships, such as entailment and contradiction, to evaluate given triplets of (question, premise, hypothesis). \\
    
    \# TASK INSTRUCTIONS \\
    Your goal is to determine whether the Premise effectively contradicts the corresponding Hypothesis. Carefully analyze each triplet, focusing on details.
    - If the premise and the hypothesis are unrelated or lack sufficient evidence to ascertain their truthfulness, label your answer as False.
    - be vigilant in identifying cases where the premise doesn't rule out the possibility of an entity (e.g., vaccine, symptom) appearing in the hypothesis. In such cases, classify the answer as False.
    - Approach each question methodically, considering the step-by-step process outlined below. \\
    
    \# INPUT DATA \\
    Question: What does trich test for? Let's think step by step. \\
    Premise: The term "trich test" can refer to two different medical tests, depending on the context. Here are the two possibilities:
    Trichomoniasis Test: Trichomoniasis is a sexually transmitted infection (STI) caused by the parasite Trichomonas vaginalis.
    The trichomoniasis test, also known as a trich test or trichomonas test, is used to detect the presence of this parasite in the body. The test is typically performed on a sample of vaginal discharge in women or urine in men. 
    Trichogram: A trichogram is a diagnostic test used to evaluate hair loss and assess the health and condition of hair follicles. It involves plucking a small number of hairs from the scalp and examining them under a microscope.
    It's important to note that without additional context, it's difficult to determine which specific test you are referring to.
    Hypothesis: Trichamoniasis- a parasitic infection that can cause your symptoms. \\
    Answer: According to the premise "trich test" refer to two different medical tests. A Trichamoniasis test is one of them, which is used to detect this parasite's presence.
    As stated in the hypothesis, the trich test is used to diagnose parasitic infections. Ths premise entails the hypothesis. The answer is False. \\
    \#\#\# \\
    Question: Can diabetics eat sweets? Let's think step by step. \\
    Premise: Individuals with diabetes are recommended to limit their consumption of sweets to one or two times per week.
    It is also suggested being selective with desserts and to focus on foods with a low glycemic index, such as high fiber foods like whole grains and legumes, as well as certain lower sugar fruits like berries, melons, and apples. \\
    Hypothesis: It is recommended that diabetics avoid sweets. \\
    Answer: The premise suggests that diabetics can eat sweets but limit their consumption. According to the hypothesis diabetics should avoid sweets.
    Diabetics are allowed to consume sweets according to the premise, but they are prohibited according to the hypothesis. There is a contradiction between the premise and the hypothesis. The answer is True. \\
    \#\#\# \\
    Question: 25 yo female with right lower abdominal pain, what might be causing it? Let's think step by step. \\
    Premise: Right lower abdominal pain in a 25-year-old female could be caused by a variety of medical conditions. \\
    Some potential causes include: Ovarian cyst: a fluid-filled sac on the ovary - Ectopic pregnancy: a pregnancy that occurs outside the uterus. \\
    Hypothesis: possible cause for right lower abdominal pain in a young female can be Appendicitis. \\
    Answer: The premise lists several potential causes of right lower abdominal pain in a 25-year-old female, not including appendicitis. The hypothesis states that Appendicitis could be a cause of right lower abdominal pain in a young female.
    There is no direct contradiction between the premise and the hypothesis, as the premise does not exclude the possibility of appendicitis as the cause of the pain. The answer is False. \\
    \#\#\# \\
    Question: Can a headache last longer than a few days? Let's think step by step. \\
    Premise: Yes, it is possible. If you are experiencing a headache that lasts longer than a few days, it is important to see a doctor to get the appropriate treatment. \\
    This will help to relieve the pain and prevent any further complications. \\
    Hypothesis: It is not a cause for concern if a headache lasts longer than a few days. \\
    Answer: This premise acknowledges that a headache can last for several days, but emphasizes that seeing a doctor to prevent further complications is important. According to this hypothesis, headaches lasting longer than a few days are not cause of concern.
    There is a contradiction between the premise and hypothesis due to the premise emphasizing the importance of seeking medical consultation, while the hypothesis posits that there is no cause for concern. The answer is True. \\
    \#\#\#\\
    
    Question: \{question\} Let's think step by step. \\
    Premise: \{llm\_answer\} \\
    Hypothesis: \{physician\_answer\} \\
    Answer:
  }
}

\clearpage

\subsection{Question-Answering}
\subsubsection*{{vanilla prompt}}
\framebox{
  \scriptsize
  \ttfamily
  \parbox{\textwidth}{
    Question: \{question\} \\
    Answer: 
  }
} 
\subsubsection*{{vanilla prompt for RAG}} 
\framebox{
  \scriptsize
  \ttfamily
  \parbox{\textwidth}{
    \# TASK: \\
    Answering question. \\
    
    \# INSTRUCTIONS: \\
    - Use ONLY our verified sources to answer the question. \\
    - If the answer is not in our sources, answer ONLY "I don't know". \\ 
    
    \# VERIFIED SOURCES \\
    Our verified sources: \{context\} \\

    \# INPUT \\
    Question: \{question\} \\

    \# OUTPUT \\
    Answer: 
  }
}
\subsubsection*{engineered prompt}
\framebox{
  \scriptsize
  \ttfamily
  \parbox{\textwidth}{
    \# TASK: 
    
    You are a medical expert and your task is to write a helpful human-like text and answer to the point. \\
    
    \# INSTRUCTIONS: 
    
    - Identify and address the patient’s focus of interest, but include relevant information even if the patient didn’t directly ask it.
    
    - If relevant, provide information about specific cases, including children, pregnant women, allergies, or chronic conditions.
    
    - If you don't know the answer, answer ONLY 'I don't know'.
    
    - If the answer contains enumeration of factors/symptoms/conditions use bullet points to organize the information. \\
    
    \# EXAMPLES: 
    
    Question: Is it safe to take Macrobid?
    
    Answer: Macrobid is generally considered safe to take when prescribed by a doctor and used according to their instructions.
    
    Rationale: The answer is good, but it is missing important information regarding special cases, such as pregnant women, kids and comorbidities.
    
    A better answer would include that Macrobid is not recommended for use in the last 2 to 4 weeks of pregnancy
    
    \#\#\#
    
    Question: Ok Do I need any antibiotics or something for yeast infection?
    
    Answer: Over-the-counter antifungal creams or suppositories applied internally to the vagina can effectively treat vaginal yeast infections, while a single oral dose of prescription fluconazole (Diflucan) may also be an option.
    
    Rationale: the answer is good however it does not include relevant explanation about the condition such as Yeast infections are commonly caused by fungi, particularly Candida albicans, which is a natural inhabitant of the body but can cause infections in different areas such as the skin, mouth, and vagina. In addition, it assumes the patient is asking about vaginal yeast infection and excludes information about other possible infections caused by yeast.
    
    \#\#\#
    
    Question: the vomiting, tight abdomen pain and burning pain may just be result of pcos?
    
    Vomiting and tight abdominal pain are not common clinical presentations of PCOS. Vomiting and abdominal pain are nonspecific symptoms which could be an indication for a variety of illnesses. In case of this presentation, you should be evaluated by a physician, and treated accordingly.
    
    Rationale: the answer is good, but it does not emphasize enough that the patient's symptoms are most likely caused by something else and what is the differential diagnosis in this case. \\
    
    \# INPUT 
    
    Question: \{question\}
    
    Answer:
  }
}
\clearpage
\subsubsection*{engineered prompt for RAG}
\framebox{
  \scriptsize
  \ttfamily
  \parbox{\textwidth}{
    \# TASK:
    
    You are a medical expert whose task is to provide a helpful, concise, human-like response to a patient's question based on verified medical sources. \\
    
    \# INSTRUCTIONS:
    
    - Use our verified sources to answer the question at the end.
    
    - If the answer is not in our sources, answer ONLY 'I don't know'.
    
    - It is important to assume that there may be a medical concern underlying the patient's questions.
    
    - If relevant, provide information about specific cases, including children, pregnant women, allergies, or chronic conditions.
    
    - If the answer contains enumeration of factors/symptoms/conditions use bullet points to organize the information.
    
    - Avoid repetition of details.\\
    
    \# EXAMPLES
    
    Question: Is it safe to take Macrobid?
    
    Answer: Macrobid is generally considered safe to take when prescribed by a doctor and used according to their instructions.
    
    Rationale: The answer is good, but it is missing important information regarding special cases, such as pregnant women, kids and comorbidities.
    
    A better answer would include that Macrobid is not recommended for use in the last 2 to 4 weeks of pregnancy
    
    \#\#\#
    
    Question: I had my gallbladder removed in 2015, so that rules out gallstones, right?
    
    Answer: Yes, as gallstones are stones that form inside the gallbladder, if the gallbladder is removed, gallstones cannot be the cause of future symptoms.
    
    Rationale: The answer is correct however it does not provide the patient with an important clinical tie as even without a gallbladder, stones can develop anywhere in the biliary system and cause similar symptoms to gallstone disease.
    
    \#\#\#
    
    Question: the vomiting, tight abdomen pain and burning pain may just be result of pcos?
    
    Answer: Polycystic ovary syndrome (PCOS) is an endocrine (hormonal) insufficiency which is characterized by polycystic ovaries and has a variety of effects including anovulation and irregular menstrual cycles which in turn could causes fertility issues, hyperandrogenism (high levels of androgens causing coarse body hair growth in a male pattern and acne) and insulin resistance causing type 2 diabetes mellitus, obesity, and hypertension.
    
    Vomiting and tight abdominal pain are not common clinical presentations of PCOS. Vomiting and abdominal pain are nonspecific symptoms which could be an indication for a variety of illnesses. In case of this presentation, you should be evaluated by a physician, and treated accordingly.
    
    Rationale: The answer is good, but it does not emphasize enough that the patient's symptoms are most likely caused by something else and what is the differential diagnosis in this case.\\
    
    \# CONTEXT
    
    Our verified sources: \{context\}\\
    
    \# INPUT 
    
    Question: \{question\}
    
    Answer:
  }
}



\end{document}